\documentclass[11pt,a4paper]{article}
\usepackage[hyperref]{acl2021}
\usepackage{times}
\usepackage{latexsym}

\usepackage{enumitem}

\usepackage{caption}
\usepackage{subcaption}

\usepackage{microtype}

\aclfinalcopy 

\usepackage{url}

\usepackage{amsmath}
\usepackage{amsfonts,bm}

\newcommand{\nn}{{\text{NN}}}
\newcommand{\ns}{{\text{NS}}}

\newtheorem{prop}{Proposition}

\newcommand{\sarrow}[1][4pt]{\!\mathrel{%
   \vcenter{\hbox{\rule[-.5\fontdimen8\textfont3]{#1}{\fontdimen8\textfont3}}}%
   \mkern-4mu\hbox{\usefont{U}{lasy}{m}{n}\symbol{41}}}\!}
\makeatletter   
\newcommand{\sveryshortarrow}[1][3pt]{\mathrel{%
    \vcenter{\hbox{\rule[-.5\fontdimen8\scriptfont3]
               {\scriptratio\dimexpr#1\relax}{\fontdimen8\scriptfont3}}}%
   \mkern-4mu\hbox{\let\f@size\sf@size\usefont{U}{lasy}{m}{n}\symbol{41}}}}
\makeatother

\def\eqref#1{equation~\ref{#1}}

\def\1{\bm{1}}

\def\va{{\bm{a}}}
\def\vb{{\bm{b}}}

\def\vd{{\bm{d}}}
\def\ve{{\bm{e}}}
\def\vf{{\bm{f}}}

\def\vh{{\bm{h}}}

\def\vs{{\bm{s}}}

\def\vw{{\bm{w}}}
\def\vx{{\bm{x}}}

\def\m1{{\bm{1}}}

\def\mW{{\bm{W}}}

\DeclareMathAlphabet{\mathsfit}{\encodingdefault}{\sfdefault}{m}{sl}
\SetMathAlphabet{\mathsfit}{bold}{\encodingdefault}{\sfdefault}{bx}{n}
\newcommand{\tens}[1]{\bm{\mathsfit{#1}}}

\def\tW{{\tens{W}}}

\def\gS{{\mathcal{S}}}
\def\gT{{\mathcal{T}}}

\def\gST{{\mathcal{S}\mathcal{T}}}

\def\sC{{\mathbb{C}}}

\def\sS{{\mathbb{S}}}

\newcommand{\Ls}{\mathcal{L}}

\newcommand{\softmax}{\mathrm{softmax}}

\DeclareMathOperator*{\argmax}{arg\,max}

\DeclareMathOperator{\real}{\rm I\!R}
\DeclareMathOperator*{\pop}{pop}
\DeclareMathOperator*{\push}{push}

\usepackage{xspace}

\newcommand{\ie}{{\em i.e.,}\xspace}
\newcommand{\eg}{{\em e.g.,}\xspace}

\newcommand{\Ni}{({\em i})~}
\newcommand{\Nii}{({\em ii})~}

\usepackage{amssymb}
\usepackage{amsmath}
\usepackage{tikz}
\usepackage{algorithm}
\usepackage{algorithmic}
\renewcommand{\algorithmiccomment}[1]{\bgroup\hfill//~#1\egroup}
\usepackage[switch]{lineno}  %

\usepackage{caption}
\usepackage{subcaption}
\usepackage{booktabs}
\usepackage{comment}

\usepackage[T1]{fontenc}

\usepackage[utf8]{inputenc}

\usepackage{microtype}

\title{A Conditional Splitting Framework for Efficient Constituency Parsing}

\author{Thanh-Tung Nguyen$^{\dagger}$$^\P$, Xuan-Phi Nguyen$^{\dagger}$$^\P$, Shafiq Joty$^\P$$^\S$, Xiaoli Li$^\dagger$$^\P$\\
  $^\P$Nanyang Technological University \\
  $^\S$Salesforce Research Asia \\
  $^\dagger$Institute for Infocomm Research, A-STAR \\
  Singapore \\
  \texttt{\{ng0155ng@e.;nguyenxu002@e.;srjoty@\}ntu.edu.sg} 
  \\
  \texttt{xlli@i2r.a-star.edu.sg} 
}

\begin{document}
\maketitle
\begin{abstract}

We introduce a generic seq2seq parsing framework that casts constituency parsing problems (syntactic and discourse parsing) into a series of conditional splitting decisions. Our parsing model estimates the conditional probability distribution of possible splitting points in a given text span and supports efficient top-down decoding, which is linear in number of nodes. The conditional splitting formulation together with efficient beam search inference facilitate structural consistency without relying on expensive structured inference. Crucially, for discourse analysis we show that in our formulation, discourse segmentation can be framed as a special case of parsing which allows us to perform discourse parsing without requiring  segmentation as a pre-requisite. Experiments show that our model achieves good results on the standard syntactic parsing tasks under settings with/without pre-trained representations and rivals state-of-the-art (SoTA) methods that are more computationally expensive than ours. In discourse parsing, our method outperforms SoTA by a good margin.   

\end{abstract}
\section{Introduction}
\begin{figure*}[t!]
\centering
\begin{subfigure}[h]{0.43\textwidth}
\scalebox{0.85}{
\begin{tikzpicture}
\node[align=left, above] at (-1,0)
{\includegraphics[width=0.9\textwidth]{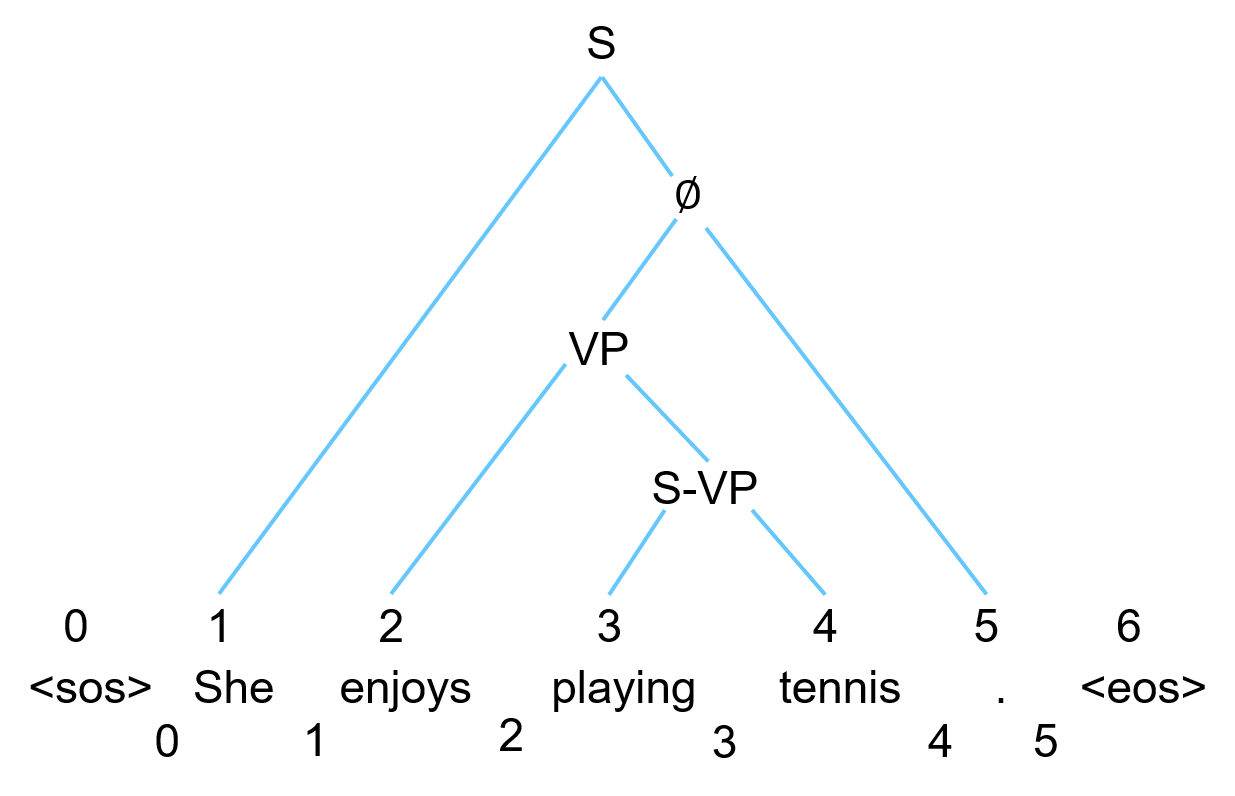}};
\small
\node[align=left, above] at (-2.7,-0.2) {\textbf{Labeled span representation}};
\node[align =left, above] at (-0.3,-0.7) {\textbf{$\sS(T) =$} \{((1, 5), S), ((2, 5), $\varnothing$), ((2, 4), VP), ((3, 4), S-VP)\}};               
\node[align=left, above] at (-1.9,-1.2) {\textbf{Boundary-based splitting representation}};         
\node[align =left, above] at (-1.0,-1.7) {\textbf{$\sC (T) =$} \{$(0,5) \sarrow 1$, $(1,5)\sarrow 4$, $(1,4)\sarrow 2$, $(2,4)\sarrow 3$\}};
\normalsize
\end{tikzpicture}
}
\label{fig:ConstituencyTree2SplittingFormat:Token-Boundary_Convert}
\end{subfigure}
\hfill 
\begin{subfigure}[h]{0.54\textwidth}
\scalebox{0.83}{
\begin{tikzpicture}
\node[align=left, above] at (-1.7,-2.9)
{\includegraphics[width=1.30\textwidth]{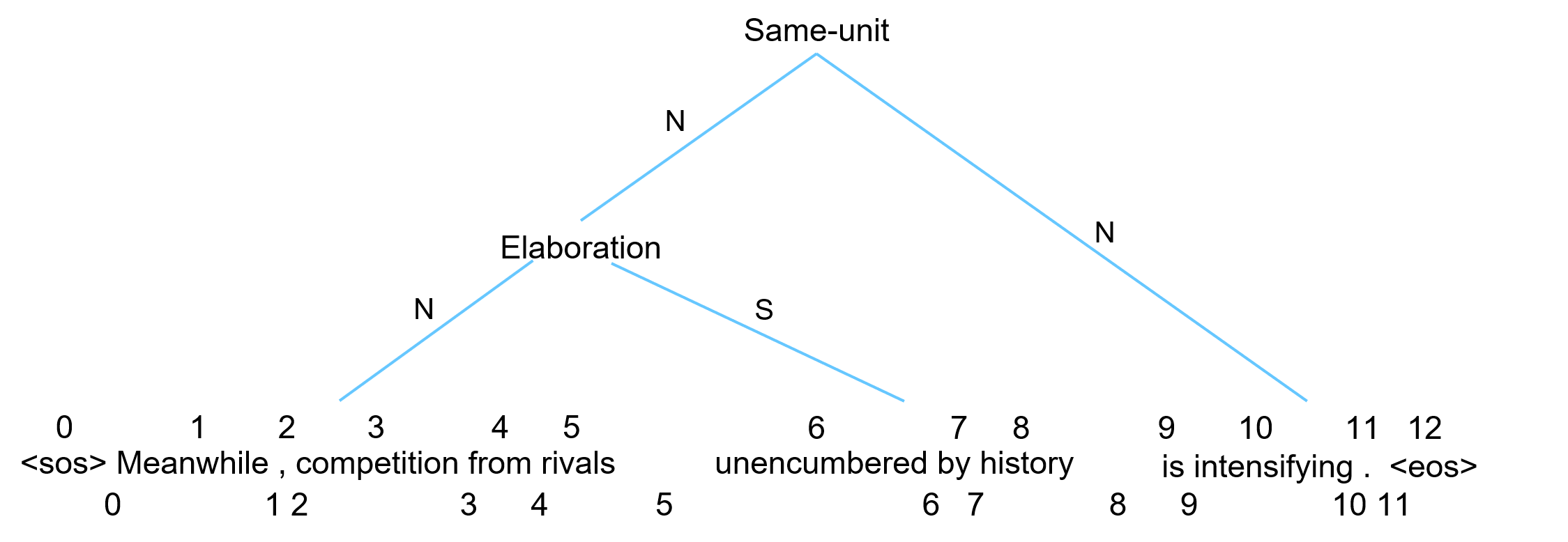}};
\small
\node[align=left, above] at (-4.8,-3.7) {\textbf{Labeled span representation}};         
\node[align =left, above] at (-2.1,-4.2) {\textbf{$\sS(DT)=$} \{((1, 8, 11), Same-Unit$_{\nn}$), ((1, 5, 8), Elaboration$_{\ns}$)\}};
\node[align=left, above] at (-4.0,-4.7) {\textbf{Boundary-based splitting representation}};         
\node[align =left, above] at (-1.8,-5.2) {\textbf{$\sC(DT) =$} \{$(0,11)\sarrow 8$, $(0,8)\sarrow 5$, $\mathbf{(0,5)\sarrow 5}$, $\mathbf{(5,8)\sarrow 8}$, $\mathbf{(8,11) \sarrow 11}$\}};
\normalsize
\end{tikzpicture}
}
\label{fig:Discourse2SplittingFormat}
\end{subfigure}
\vspace{-1.5em}
\caption{\small A syntactic  tree at the left and a discourse tree (DT) at the right; both have a constituency structure. The internal nodes in the discourse tree (\emph{Elaboration}, \emph{Same-Unit}) represent coherence relations and the edge labels indicate the nuclearity statuses (`N' for Nucleus and `S' for Satellite) of the child spans.
Below the tree, we show the labeled span and splitting representations. The bold splits in the DT representation ($\sC(DT)$) indicate the end of further splitting into smaller spans (\ie\ they are EDUs).}
%
\label{fig:ConstituencyDiscourse2SplittingFormat}
\end{figure*}

A number of  formalisms have been introduced to analyze natural language at different linguistic levels. This includes syntactic structures in the form of phrasal and  dependency trees, semantic structures in the form of meaning
representations \cite{banarescu-etal-2013-abstract,artzi-etal-2013-semantic}, and discourse structures with Rhetorical Structure Theory (RST) \cite{Mann88} or Discourse-LTAG \cite{Webber04}. Many of these formalisms have a \emph{constituency} structure, where textual units (\eg\ phrases, sentences) are organized into nested constituents. For example, Figure \ref{fig:ConstituencyDiscourse2SplittingFormat} shows examples of a phrase structure tree and a sentence-level discourse tree (RST) that respectively represent how the phrases and clauses are hierarchically organized into a constituency structure. Developing efficient and effective parsing solutions has always been a key focus in NLP. In this work, we consider both phrasal (syntactic) and discourse parsing.

In recent years, neural end-to-end parsing methods have outperformed traditional methods that use grammar, lexicon and hand-crafted features. These methods can be broadly categorized based on whether they employ a greedy transition-based, a globally optimized chart parsing or  a greedy top-down algorithm. Transition-based parsers \citep{dyer-etal-2016-recurrent,cross-huang-2016-span,liu-zhang-2017-shift,Wang-acl-2017} generate trees auto-regressively as a form of shift-reduce decisions. Though computationally attractive, the local decisions made at each step may propagate errors to subsequent steps due to exposure bias \cite{NIPS2015_e995f98d}. Moreover, there may be mismatches in shift and reduce steps, resulting in invalid trees. 

Chart based methods, on the other hand, train neural scoring functions to model the tree structure globally \citep{durrett-klein-2015-neural,gaddy-etal-2018-whats,kitaev-klein-2018-constituency,ijcai2020-560,joty-etal-2012-novel,joty-etal-2013-combining}. By utilizing dynamic programming, these methods can perform  exact inference to combine these constituent scores into finding the highest probable tree. However, they are generally slow with at least $\mathcal{O}(n^3)$ time complexity. Greedy top-down parsers find the split points recursively and have received much attention lately due to their efficiency, which is usually $\mathcal{O}(n^2)$ \citep{minimal-span-based-parsing, shen-etal-2018-straight,lin-etal-2019-unified, nguyen-etal-2020-efficient}. However, they still suffer from exposure bias, where one incorrect splitting step may affect subsequent steps.

Discourse parsing in RST requires an additional step -- \emph{discourse segmentation} which involves breaking the text into contiguous clause-like units called Elementary Discourse Units or EDUs (Figure \ref{fig:ConstituencyDiscourse2SplittingFormat}). Traditionally, segmentation has been considered separately and as a prerequisite step for the parsing task which links the EDUs (and larger spans) into a discourse tree \citep{soricut-marcu-2003-sentence,joty-etal-2012-novel, Wang-acl-2017}. In this way, the errors in discourse segmentation can propagate to discourse parsing \cite{lin-etal-2019-unified}.

In this paper, we propose a generic top-down neural framework for constituency parsing that we validate on both syntactic and sentence-level discourse parsing. Our main contributions are:

\begin{itemize}[leftmargin=*,itemsep=0em]
\item We cast the constituency parsing task into a series of conditional splitting decisions and use a seq2seq architecture to model the splitting decision at each decoding step. Our parsing model, which is an instance of a Pointer Network \citep{VinyalsNIPS2015}, estimates the pointing score from a span to a splitting boundary point, representing the likelihood that the span will be split at that point and create two child spans. 

\item The conditional probabilities of the splitting decisions are optimized using a cross entropy loss and structural consistency is maintained through {a global pointing mechanism}. The training process can be fully parallelized without requiring  structured inference as in \cite{shen-etal-2018-straight,gomez-rodriguez-vilares-2018-constituent, nguyen-etal-2020-efficient}. 

\item Our model enables efficient top-down decoding with $\mathcal{O}(n)$ running time like transition-based parsers, while also supporting a customized beam search to get the best tree by searching through a reasonable search space of high scoring trees. The beam-search inference along with the structural consistency from the modeling makes our approach competitive with existing structured chart methods for syntactic \cite{kitaev-klein-2018-constituency} and discourse parsing \cite{ijcai2020-560}. Moreover, our parser does not rely on any handcrafted features (not even part-of-speech tags), which makes it more efficient and be flexible to different domains or languages.

\item For discourse analysis, we demonstrate that our method can effectively find the segments (EDUs) by simply performing one additional step in the top-down parsing process. In other words, our method can parse a text into the discourse tree without needing discourse segmentation as a pre-requisite; instead, it produces the segments as a by-product. To the best of our knowledge, this is the first model that can perform segmentation and parsing in a single embedded framework. 

\end{itemize}

In the experiments with English Penn Treebank, our model without pre-trained representations achieves 93.8 F1, outperforming all existing methods with similar time complexity. With pre-training, our model pushes the F1 score to 95.7, which is on par with the SoTA while supporting faster decoding with a speed of over 1,100 sentences per second (fastest so far). {Our model also performs competitively with SoTA methods on the multilingual parsing tasks in the SPMRL 2013/2014 shared tasks}. In discourse parsing, our method establishes a new SoTA in end-to-end sentence-level parsing performance on the RST Discourse Treebank with an F1 score of 78.82.

We make our code available at \href{https://ntunlpsg.github.io/project/condition-constituency-style-parser/}{https://ntunlpsg.github.io/project/condition-constituency-style-parser/}

\section{Parsing as a Splitting Problem}

Constituency parsing (both syntactic and discourse) can be considered as the problem of finding a set of labeled spans over the input text \citep{minimal-span-based-parsing}. 
Let $\sS(T)$ denote the set of \textit{labeled  spans} for a parse tree $T$, which can formally be expressed as (excluding the trivial singleton span layer):    
\begin{equation}
\sS(T) := \{ ((i_t, j_t), l_t) \}_{t=1}^{|\sS (T)|} \text{ for } {i_t < j_t} 
\end{equation}
where $l_t$ is the label of the text span $(i_t, j_t)$ encompassing tokens from index $i_t$ to index $j_t$. 

Previous approaches to syntactic parsing \citep{minimal-span-based-parsing, kitaev-klein-2018-constituency, nguyen-etal-2020-efficient} train a neural model to score each possible span and then apply a greedy or dynamic programming algorithm  to find the parse tree. In other words, these methods are span-based  formulation. 

In contrary, we formulate constituency parsing as the problem of finding the splitting points in a recursive, top-down manner. For each parent node in a tree that spans over $(i,j)$, our parsing model is trained to point to the \emph{boundary} between the tokens at $k$ and $k+1$ positions to split the parent span into two child spans $(i,k)$ and $(k+1,j)$. This is done through the Pointing mechanism \citep{VinyalsNIPS2015}, where each splitting decision is modeled as a multinomial distribution over the input elements, which in our case are the token boundaries.

The correspondence between token- and boundary-based representations of a tree is straight-forward. After including the start ($<${sos}$>$) and end ($<${eos}$>$) tokens, the token-based span $(i,j)$ is equivalent to the boundary-based span $(i-1,j)$ and the boundary between $i$-{th} and $(i+1)$-th tokens is indexed as $i$. For example, the (boundary-based) span ``enjoys playing tennis'' in  Figure \ref{fig:ConstituencyDiscourse2SplittingFormat} is defined as $(1,4)$. Similarly, the boundary between the tokens ``enjoys'' and ``playing'' is indexed with  $2$.\footnote{We use the same example from \cite{minimal-span-based-parsing,shen-etal-2018-straight,nguyen-etal-2020-efficient} to distinguish the differences between the methods.}

Following the common practice in syntactic parsing, we binarize the $n$-ary tree by introducing a dummy label $\varnothing$.  
 We also collapsed the nested labeled spans in the unary chains into unique atomic labels, such as S-VP in Figure~\ref{fig:ConstituencyDiscourse2SplittingFormat}.
Every span represents an internal node in the tree, which has a left and a right child. 
Therefore, we can represent each internal node by its split into left and right children. Based on this, we define the set of splitting decisions $\sC (T)$ for a syntactic tree $T$ as follows.
\begin{prop}
\label{prop:1} 
A binary syntactic tree $T$ of a sentence containing $n$ tokens can be transformed into a set of splitting decisions $\sC(T) = \{(i,j)\sarrow  k: i < k < j\}$ such that the parent span $(i,j)$ is split into two child spans $(i,k)$ and $(k,j)$.
\end{prop}
An example of the splitting representation of a tree is shown in Figure \ref{fig:ConstituencyDiscourse2SplittingFormat} (without the node labels).  
Note that our transformed representation has a one-to-one mapping with the tree since each splitting decision corresponds to one and only one internal node in the tree. 
We follow a depth-first order of the decision sequence, which in our preliminary experiments showed more consistent performance than other alternatives like breadth-first order.

\paragraph{Extension to End-to-End Discourse Parsing}

Note that in syntactic parsing, the split position must be within the span but not at its edge, that is, $k$ must satisfy $i < k < j$ for each boundary span $(i,j)$. Otherwise, it will not produce valid sub-trees. In this case, we keep splitting until each span contains a single leaf token. However, for discourse trees, each leaf is an EDU -- a clause-like unit that can contain one or multiple tokens. 

Unlike previous studies which assume discourse segmentation as a pre-processing step, we propose a unified formulation that treats segmentation as one additional step in the top-down parsing process. To accommodate this, we relax Proposition 1 as:

\begin{prop}
\label{prop:2} 
A binary \underline{discourse} tree $DT$ of a text containing $n$ tokens can be transformed into a set of splitting decisions $\sC(DT) = \{(i,j)\sarrow  k: i < k \leq j\}$ such that the parent span $(i,j)$ gets split into two child spans $(i,k)$ and $(k,j)$ for $k<j$ {or a terminal span or EDU for $k=j$ (end of splitting the span further)}.
\end{prop}

We illustrate it with the DT example in Figure \ref{fig:ConstituencyDiscourse2SplittingFormat}. Each splitting decision in $\sC(DT)$ represents either the splitting of the parent span into two child spans (when the splitting point is strictly within the span) or the end of any further splitting (when the splitting point is the right endpoint of the span). {By making this simple relaxation, our formulation can not only generate the discourse tree (in the former case) but can also find the discourse segments (EDUs) as a by-product (in the latter case).}

\begin{figure*}[t!]
\centering
\includegraphics[width=0.7\textwidth]{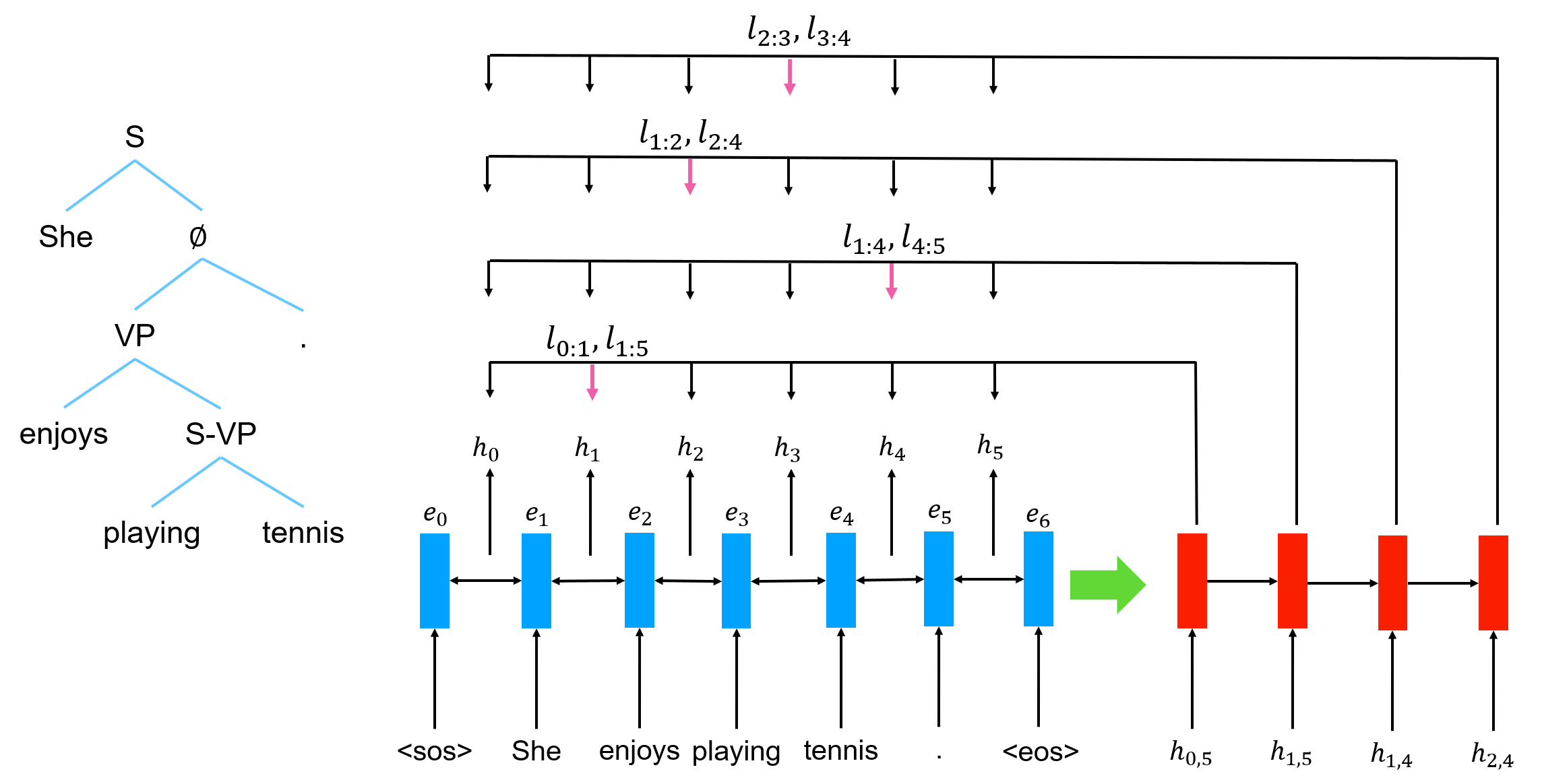}
\caption{\small Our syntatic parser along with the decoding process for a given sentence. The input to the decoder at each step is the representation of the span to be split. We predict the splitting point using a biaffine function between the corresponding decoder state and the boundary-based encoder representations. A label classifier is used to assign labels
to the left and right spans.
\normalsize}
\label{fig:constituency_parsing_whole_architecture}
\end{figure*}

\section{Seq2Seq Parsing Framework}

Let $\sC(T)$ and $L(T)$ respectively denote the structure (in split representation) and labels of a tree $T$ (syntactic or discourse) for a given text $\vx$. We can express the probability of the tree as: 

\small 
\begin{align}
\begin{split}
P_{\theta}(T|\vx) & =  P_{\theta}(L(T),\sC(T)|\vx) \\
   & =  P_{\theta}(L(T)|\sC(T),\vx) P_{\theta}(\sC(T)|\vx) 
\end{split}
\label{eq1}
\end{align}
\normalsize

\noindent  This factorization allows us to first infer the tree structure from the input text, and then find the corresponding labels. 
As discussed in the previous section, we consider the structure prediction as a sequence of splitting decisions to generate the tree in a top-down manner. Specifically, at each decoding step $t$, the output $y_t$ represents the splitting decision $(i_t,j_t) \sarrow k_t$ and $y_{<t}$ represents the previous splitting decisions. 
Thus, we can express the probability of the tree structure as follows:

\small 
\begin{align} 
\begin{split}
 \hspace{-0.5em} P_{\theta}(\sC(T)|\vx)   & =  \hspace{-1em} \prod\limits_{y_t \in \sC(T)} P_{\theta}(y_t | y_{<t},\vx) \\
 & =  \hspace{-0.5em} \prod \limits_{t=1}^{|\sC(T)|} \hspace{-0.2em} P_{\theta} ( (i_t,j_t)\sarrow k_t | ((i,j)\sarrow k)_{<t},\vx)
\end{split}
\label{eq2}
\end{align}
\normalsize

\noindent This can effectively be modeled within a Seq2Seq pointing framework as shown in Figure \ref{fig:constituency_parsing_whole_architecture}. At each step $t$, the decoder autoregressively predicts the split point $k_t$ in the input by conditioning on the current input span $(i_t,j_t)$ and previous splitting decisions $(i,j)\sarrow k)_{<t}$. This conditional splitting formulation (decision at step $t$ depends on previous steps) can help our model to find better trees compared to  \emph{non-conditional} top-down parsers \citep{minimal-span-based-parsing, shen-etal-2018-straight,nguyen-etal-2020-efficient}, thus bridging the gap between the global (but expensive) and the local (but efficient) models. The labels $L(T)$ can be modeled by using a label classifier, as described later in the next section.

\subsection{Model Architecture}

We now describe the components of our parsing model: the sentence encoder, the span representation, the pointing model and the labeling model.

\paragraph{Sentence Encoder}
Given an input sequence of $n$ tokens $\vx = (x_1, \ldots, x_n)$, we first add $<${sos}$>$ and $<${eos}$>$ markers to the  sequence. After that, each token $t$ in the sequence is mapped into its dense vector representation $\ve_t$ as 

\small 
\begin{align} 
    \ve_t = [\ve_t^{\text{char}} , \ve_t^{\text{word}} ]
\end{align}
\normalsize

\noindent where $\ve_t^\text{char}$, $\ve_t^\text{word}$ are respectively the character and word embeddings of token $t$. Similar to \cite{kitaev-klein-2018-constituency, nguyen-etal-2020-efficient}, we use a character LSTM to compute the character embedding of a token. We experiment with both randomly initialized and pretrained token embeddings. When pretrained embedding is used, the character embedding is replaced by the pretrained token embedding. The token representations are then passed to a 3-layer Bi-LSTM encoder to obtain their contextual representations. In the experiments, we find that even without the POS-tags, our model performs competitively with other baselines that use them.

\begin{figure}[t!]
\centering
\includegraphics[width=0.35\textwidth]{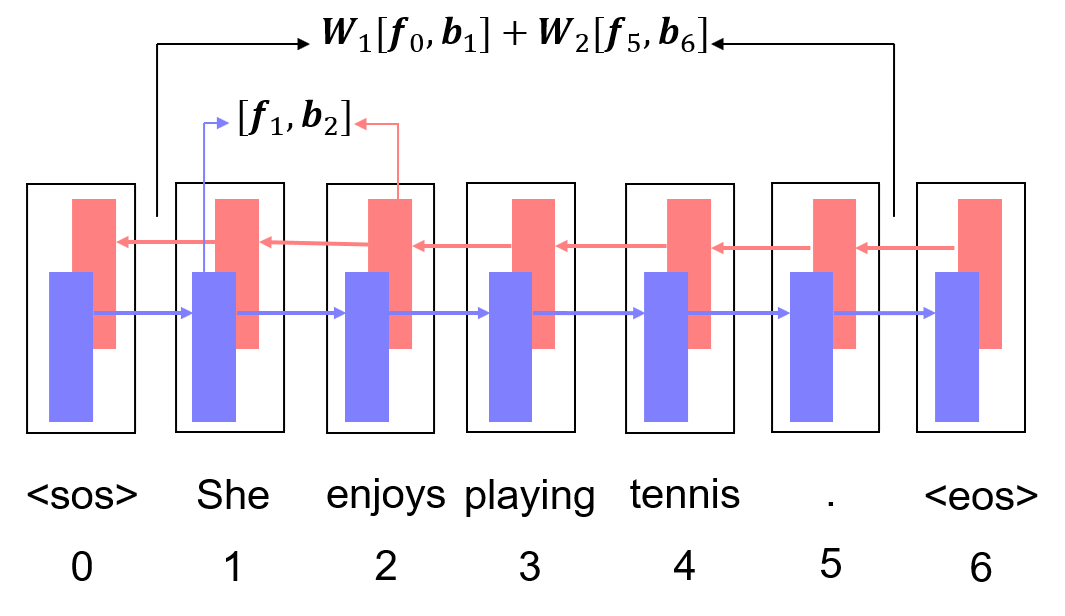}
\caption{\small Illustration of our  boundary-based span encoder. Here we have shown the representation for the boundary at $1$ and the representation of the boundary-based span $(0, 5)$ that corresponds to the sentence \emph{``She enjoys playing tennis .''}.}
\label{fig:boundary_span_representation}
\end{figure}

\paragraph{Boundary and Span Representations}
To represent each boundary between positions $k$ and $k+1$, we use the fencepost representation \citep{cross-huang-2016-span, minimal-span-based-parsing}:

\small 
\begin{align}
    \vh_{k} = [\vf_{k} , \vb_{k+1}]
\end{align}
\normalsize

\noindent where $\vf_k$ and $\vb_{k+1}$ are the forward and backward LSTM hidden vectors at positions $k$ and $k+1$, respectively. To represent the span $(i,j)$, we compute a linear combination of the two endpoints

\small 
\begin{align}
    \vh_{i,j} = \mW_1 \vh_i + \mW_2 \vh_j
\end{align}
\normalsize

This span representation will be used as input to the decoder. Figure \ref{fig:boundary_span_representation} shows the boundary-based span representations for our example.

\paragraph{The Decoder}

Our model uses a unidirectional LSTM as the decoder. At each decoding step $t$, the decoder takes as input the corresponding span $(i,j)$ (specifically, $\vh_{i,j}$)  and its previous state $\vd_{t-1}$ to generate the current state $\vd_t$ and then apply a biaffine function \citep{DozatM17} between $\vd_t$ and \textbf{all} of the encoded boundary representations $(\vh_0, \vh_1, \ldots, \vh_n)$ as follows:

\small 
\begin{align}
\vd'_t =  \text{MLP}_d(\vd_t) \hspace{1em} \vh'_i = \text{MLP}_h(\vh_i) \\
s_{t,i} = {\vd'_t}^T \mW_{dh} \vh'_i + {\vh'_i}^T\vw_h\\
a_{t,i} = \frac{\exp(s_{t,i})}{\sum_{i=1}^n \exp(s_{t,i})}
\label{eq:decoder_pointing}
\end{align}
\normalsize

\noindent where each $\text{MLP}$ operation includes a linear transformation with LeakyReLU activation to transform $\vd$ and $\vh$ into equal-sized vectors, and $\mW_{dh} \in \real^{d  \times d}$ and $\vw_h \in \real^d$ are respectively the weight matrix and weight vector for the biaffine function. The biaffine scores are then passed through a softmax layer to acquire the pointing distribution $\va_t \in [0,1]^n $ for the splitting decision.

When decoding the tree during inference,  at each step we only examine the `valid' splitting points between $i$ and $j$ -- for syntactic parsing, it is $i<k<j$ and for discourse parsing, it is $i<k \leq j$.

\paragraph{Label Classifier}

For syntactic parsing, we perform the label assignments for a span $(i,j)$ as:

\small 
\begin{align}
\vh^l_i =  \text{MLP}_l(\vh_i); \hspace{1em} \vh^r_j = \text{MLP}_r(\vh_j) \\
P_{\theta}(l|i,j) = \softmax ((\vh^l_i)^T \tW_{lr} \vh^r_j \nonumber \\
+ (\vh^l_i)^T\mW_l + (\vh^r_j)^T\mW_r + \vb) \\
l_{i,j}= \argmax_{l \in L}P_{\theta}(l|i,j)
\label{eq:constituency_labelling}
\end{align}
\normalsize

\noindent where each of $\text{MLP}_l$ and $\text{MLP}_r$ includes a linear transformation with LeakyReLU activations to transform the left and right spans into equal-sized vectors, and $\tW_{lr}\in \real^{d \times L \times d}, \mW_l \in \real^{d \times L},\mW_r\in \real^{d \times L}$ are the weights and $\vb$ is a bias vector with $L$ being the number of phrasal labels.

For discourse parsing, we perform label assignment \emph{after} every split decision since the label here represents the relation between the child spans. Specifically, as we split a span $(i,j)$ into two child spans $(i,k)$ and $(k,j)$, we determine the relation label as the following.

\small 
\begin{align}
\vh^l_{ik} =  \text{MLP}_l([\vh_i,\vh_k]); \hspace{1em} \vh^r_{kj} = \text{MLP}_r([\vh_k,\vh_j]) \\
P_{\theta}(l|(i,k),(k,j)) = \softmax ((\vh^l_{ik})^T \tW_{lr} \vh^r_{kj} \nonumber \\
+ (\vh^l_{ik})^T\mW_l + (\vh^r_{kj})^T\mW_r + \vb)\\
l_{(i,k),(k,j)}= \argmax_{l \in L}P_{\theta}(l|(i,k),(k,j))
\label{eq:discourse_labelling}
\end{align}
\normalsize
where $\text{MLP}_l, \text{MLP}_r$, $\tW_{lr}, \mW_l,\mW_r, \vb$ are similarly defined.

\paragraph{Training Objective}

The total loss is simply the sum of the cross entropy losses for predicting the structure (split decisions) and the labels:


\small 
\begin{align}
\Ls_{\text{total}} (\theta) = \Ls_{\text{split}} (\theta_e,\theta_d) + 
\Ls_{\text{label}} (\theta_e, \theta_{\text{label}})
\end{align}
\normalsize

\noindent where $\theta = \{\theta_e, \theta_d, \theta_{\text{label}}\}$ denotes the overall model parameters, which includes the encoder parameters $\theta_e$ shared by all components, parameters for splitting $\theta_{d}$ and parameters for labeling $\theta_{\text{label}}$.

\subsection{Top-Down Beam-Search Inference}

As mentioned, existing top-down syntactic parsers do not consider the decoding history. They also perform greedy inference. With our conditional splitting formulation, our method can not only model the splitting history but also enhance the search space of high scoring trees through beam search.

At each step, our decoder points to \emph{all} the encoded \emph{boundary} representations which ensures that the pointing scores are in the same scale, allowing a fair comparison between the total scores of all candidate subtrees. With these uniform scores, we could apply a beam search to infer the most probable tree using our model. Specifically, the method generates the tree in depth-first order while maintaining top-$B$ (beam size) partial trees at each step. It terminates exactly after $n-1$ steps, which matches the number of internal  nodes in the tree. Because beam size $B$ is constant with regards to the sequence length, we can omit it in the Big $\mathcal{O}$ notation. Therefore, each decoding step with beam search can be parallelized ($\mathcal{O}(1)$ complexity) using GPUs. This makes our algorithm run at $\mathcal{O}(n)$ time complexity, which is faster than most top-down methods.  If we strictly use CPU, our method runs at $\mathcal{O}(n^2)$, while chart-based parsers run at $\mathcal{O}(n^3)$. Algorithm \ref{alg1} illustrate the syntactic tree inference procedure. We also propose a similar version of the inference algorithm for discourse parsing in the Appendix.

\begin{algorithm}[ht!]
\footnotesize
    \caption{Syntactic Tree Inference with Beam Search}
    \label{alg1}
    \begin{algorithmic}[1]
    \REQUIRE Sentence length $n$; beam width $B$; boundary-based encoder states: $(\vh_0, \vh_1, \ldots, \vh_n)$; label scores: $P_{\theta} (l|i,j)$, $0\leq i < j \leq n, l \in \{1, \ldots, L\}$, initial decoder state $\vs$.
    \ENSURE Parse tree $T$
    \STATE $L_d = n-1$ \algorithmiccomment{Decoding length}
    \STATE {beam} = array of $L_d$ items            \algorithmiccomment{List of empty beam items}
    \STATE init\_tree$=[(0,n),(0,0),\ldots, (0,0)]$ \algorithmiccomment{$n-2$ paddings (0,0)}
    \STATE beam{[0]} = $(0, \vs, \text{init\_tree})$ \algorithmiccomment{Init 1st item(log-prob,state,tree)}
    \FOR {$t=1\ \TO\ L_d$}
        \FOR {$(\text{logp}, \vs, \text{tree}) \in \text{beam}[t-1]$} 
            \STATE $(i,j) =\text{tree}[t-1]$ \algorithmiccomment{Current span to split}
            \STATE $\va, \vs'= \text{decoder-step}(\vs, \vh_{i,j})$ \algorithmiccomment{$\va$: split prob. dist.}
            \FOR {$(k, p_k) \in \text{top-}B(\va)\ \AND\ i<k<j $}
                \STATE $\text{curr-tree}=\text{tree}$
                \IF{$k > i+1$}    
                    \STATE $\text{curr-tree}[t]=(i,k)$
                \ENDIF
                \IF{$j > k+1$}    
                    \STATE $\text{curr-tree}[t+j-k-1]=(k,j)$
                \ENDIF
                \STATE push ({logp} + $\log(p_k), \vs',$ {curr-tree}) to beam[t]
            \ENDFOR
        \ENDFOR
        \STATE prune beam[t] \algorithmiccomment{Keep top-$B$ highest score trees}
    \ENDFOR
    \STATE $\text{logp*},\vs^*,\gS^* = \argmax_{\text{logp}} \text{beam}[L_d]$ \algorithmiccomment{$\gS^*$: best structure}
    \STATE $\text{labeled-spans} =[(i,j, \argmax_{l} P_{\theta} (l|i,j)) ~~\forall (i,j) \in \gS^*]$
    \STATE $\text{labeled-singletons} =[(i,i+1, \argmax_{l}P_{\theta} (l|i,i+1)) ~\text{for } i = \{0, \ldots, n-1 \} ]$
    \STATE $\gT=\text{labeled-spans}  \cup \text{labeled-singletons}$
  \end{algorithmic}
\end{algorithm}

By enabling beam search, our method can find the best tree by comparing high scoring trees within a reasonable search space, making our model competitive with existing  structured (globally) inference methods that use more expensive algorithms like CKY and/or larger models \citep{kitaev-klein-2018-constituency,ijcai2020-560}.

\section{Experiment}
\paragraph{Datasets and Metrics} 

To show the effectiveness of our approach, we conduct experiments on both syntactic and sentence-level RST parsing tasks.\footnote{Extending the discourse parser to the document level may require handling of intra- and multi-sentential constituents differently, which we leave for future work.} We use the standard Wall Street Journal (WSJ) part of the Penn Treebank (PTB) \cite{PTB:Marcus:1993} for syntactic parsing and RST Discourse Treebank (RST-DT) \cite{RST:DT:2002} for discourse parsing. For syntactic parsing, we also experiment with the multilingual parsing tasks on seven different languages from the SPMRL 2013-2014 shared task \citep{seddah-etal-2013-overview}: Basque, French, German, Hungarian, Korean, Polish and Swedish.

For evaluation on syntactic parsing, we report the standard labeled precision (LP), labeled recall (LR), and labelled F1 computed by \texttt{evalb}\footnote{\url{http://nlp.cs.nyu.edu/evalb/}}. For evaluation on RST-DT, we report the standard span, nuclearity label, relation label F1 scores, computed using the implementation of \cite{lin-etal-2019-unified}.\footnote{\url{https://github.com/ntunlpsg/UnifiedParser_RST}}

\subsection{English (PTB) Syntactic Parsing}

\paragraph{Setup}
We follow the standard train/valid/test split, which uses Sections 2-21 for training, Section 22 for development and Section 23 for evaluation. This results in 39,832 sentences for training, 1,700 for development, and 2,416 for testing.
For our model, we use an LSTM encoder-decoder framework with a 3-layer bidirectional encoder and 3-layer unidirectional decoder. The word embedding size is 100 while the character embedding size is 50; the LSTM hidden size is 400. The hidden dimension in MLP modules and biaffine function for split point prediction is 500. The beam width $B$ is set to 20. We use the Adam optimizer \citep{KingmaB14} with a batch size of 5000 tokens, and an initial learning rate of $0.002$ which decays at the rate $0.75$ exponentially at every 5k steps. Model selection for final evaluation is performed based on the labeled F1 score on the development set.

\paragraph{Results without Pre-training} From the results shown in Table \ref{tab:ptb_single}, we  see that our model achieves an F1 of $93.77$, the highest among models that use \emph{top-down} methods. Specifically, our parser outperforms \citet{minimal-span-based-parsing,shen-etal-2018-straight} by about $2$ points in F1-score and \citet{nguyen-etal-2020-efficient} by $\sim$$1$ point. Notably, without beam search (beam width 1 or greedy decoding), our model achieves an F1 of $93.40$, which is still better than other top-down methods. Our model also performs competitively with CKY-based methods like \cite{kitaev-klein-2018-constituency, ijcai2020-560, wei-etal-2020-span, ZhouZ19}, while these methods run slower than ours. 

Plus, \citet{ZhouZ19} uses external supervision (\emph{head} information) from the dependency parsing task. Dependency parsing models, in fact, have a strong resemblance to the pointing mechanism that our model employs \citep{ma-etal-2018-stack}. As such, integrating dependency parsing information into our model may also be beneficial. We leave this for future work.

\begin{table}[t]
\begin{center}
\resizebox{0.9\columnwidth}{!}{%
\setlength\tabcolsep{3.2pt}
\begin{tabular}{l|ccc} 
\toprule
{\bf Model}         & LR & LP & F1 \\
\midrule
\multicolumn{4}{c}{\textbf{Top-Down Inference}}    \\
\citet{minimal-span-based-parsing} & 93.20 & 90.30 & 91.80 \\
\citet{shen-etal-2018-straight}  & 92.00 & 91.70 & 91.80 \\
\citet{nguyen-etal-2020-efficient} &92.91 & 92.75  & 92.78 \\
{Our Model} & 93.90 & 93.63 & 93.77 \\
\midrule
\multicolumn{4}{c}{\textbf{CKY/Chart Inference} }  \\
\citet{gaddy-etal-2018-whats}  & 91.76 & 92.41 & 92.08 \\
\citet{kitaev-klein-2018-constituency} &93.20 &93.90&93.55\\
\citet{wei-etal-2020-span} & 93.3 & 94.1 & 93.7 \\
\citet{ijcai2020-560} &93.84 &93.58& 93.71 \\
\midrule
\multicolumn{4}{c}{\textbf{Other Approaches}}   \\
\citet{gomez-rodriguez-vilares-2018-constituent} &- &- &90.7\\
\citet{liu-zhang-2017-shift} &- &- &91.8 \\
\citet{stern-etal-2017-effective} &92.57&92.56&92.56\\
\citet{ZhouZ19} &93.64 &93.92& 93.78\\
\bottomrule
\end{tabular}
\setlength\tabcolsep{6pt}
}
\caption{\label{tab:ptb_single}Results for single models (no pre-training) on the PTB WSJ test set, Section 23.}
\label{table:ptb_single_model}
\end{center}
\end{table}

\begin{table}[t]
\begin{center}
\resizebox{0.6\columnwidth}{!}{%
\begin{tabular}{lc} 
\toprule
{\bf Model}         & {\bf F1} \\
\midrule
\citet{nguyen-etal-2020-efficient}& 95.5 \\
Our model & 95.7 \\
\midrule
\citet{kitaev-etal-2019-multilingual} &95.6\\
\citet{ijcai2020-560} &95.7\\
\citet{wei-etal-2020-span} &95.8\\
\citet{ZhouZ19} & 95.8\\
\bottomrule
\end{tabular}
}
\caption{\label{tab:ptb_pretrain}Results on PTB WSJ test set with pretraining.}
\label{table:ptb_pretrained_model}
\end{center}
\end{table}

\paragraph{Results with Pre-training} Similar to \cite{kitaev-klein-2018-constituency,kitaev-etal-2019-multilingual}, we also evaluate our parser with BERT embeddings \cite{devlin-etal-2019-bert}. They fine-tuned  \emph{Bert-large-cased} on the task, while in our work keeping it frozen was already good enough {(gives training efficiency)}. As shown in Table \ref{table:ptb_pretrained_model}, our model achieves an F1 of $95.7$, which is on par with SoTA models. However, our parser runs faster than other methods. Specifically, our model runs at {$\mathcal{O}(n)$} time complexity, while CKY needs $\mathcal{O}(n^3)$. 
Comprehensive comparisons on parsing speed are presented later.

\subsection{SPMRL Multilingual Syntactic Parsing}

We use the identical hyper-parameters and optimizer setups as in English PTB. We follow the standard train/valid/test split provided in the SPMRL datasets; details are reported in the Table \ref{table:spmrl_corpus_stats}.

\begin{table}[h!]
\begin{center}
\resizebox{0.7\columnwidth}{!}{%
\begin{tabular}{l|ccc} 
\toprule
{\bf Language} & Train & Valid & Test \\
\midrule
{\bf Basque} & 7,577 &   948 & 946 \\
{\bf French} & 14,759 &  1,235 &  2,541 \\
{\bf German} & 40,472 & 5,000  & 5,000  \\
{\bf Hungarian} & 8,146 & 1,051 & 1,009\\
{\bf Korean} & 23,010 &  2,066  & 2,287\\ 
{\bf Polish} & 6,578  &  821 & 822 \\
{\bf Swedish} & 5,000 & 494 & 666 \\
\bottomrule
\end{tabular}
}
\caption{\label{tab:spmrl_dataset_stat}SPMRL Multilingual dataset split.}
\label{table:spmrl_corpus_stats}
\end{center}
\end{table}

From the results in Table \ref{table:spmrl_single_model}, we see that our model achieves the highest F1 in French, Hungarian and Korean and higher than the best baseline by $0.06$, $0.15$ and $0.13$, respectively. Our method also rivals existing SoTA methods on other languages even though some of them use predicted POS tags \cite{nguyen-etal-2020-efficient} or bigger models ($75$M parameters) \cite{kitaev-klein-2018-constituency}. Meanwhile, our model is smaller ($31$M), uses no extra information and runs 40\% faster.

\begin{table*}[t]
\begin{center}
\resizebox{1.6\columnwidth}{!}{%
\begin{tabular}{l|ccccccc} 
\toprule
{\bf Model}         & {\bf Basque} & {\bf French} & {\bf German}   & {\bf Hungarian} & {\bf Korean} & {\bf Polish} & {\bf Swedish}\\
\midrule
\citet{spmrl2014}$^{+}$  &88.24 &82.53 &81.66 & 91.72 &83.81 &90.50 &85.50\\
\citet{coavoux-crabbe-2017-multilingual}$^{+}$ & 88.81 & 82.49 & 85.34 & 92.34 & 86.04 & 93.64 & 84.0 \\
\citet{kitaev-klein-2018-constituency}   & 89.71 & 84.06 & 87.69  & 92.69 & 86.59 & 93.69 & 84.45 \\
\citet{nguyen-etal-2020-efficient}$^{+}$ & 90.23 & 82.20 & 84.91  & 91.07 & 85.36 & 93.99 & 86.87  \\
Our Model & 89.74 & 84.12 &85.21 & 92.84 & 86.72 & 92.10  & 85.81  \\
\bottomrule
\end{tabular}
\setlength\tabcolsep{6pt}
}
\caption{\label{tab:spmrl_dataset}Results on SPMRL test sets without pre-training. The sign $^{+}$ denotes that systems use predicted POS tags.}
\label{table:spmrl_single_model}
\end{center}
\end{table*}

\subsection{Discourse Parsing}

\paragraph{Setup}
For discourse parsing, we follow the standard split from \cite{lin-etal-2019-unified}, which has 7321 sentence-level discourse trees for training and 951 for testing. We also randomly select $10\%$ of the training {for validation}. Model selection for testing is performed based on the F1 of relation labels on the validation set. We use the same model settings as the constituency parsing experiments, with BERT as pretrained embeddings.\footnote{ \citet{lin-etal-2019-unified} used ELMo \cite{Peters:2018} as pretrained embeddings. With BERT, their model performs worse which we have confirmed with the authors.}

\paragraph{Results} Table \ref{table:parseg-results} compares the results on the discourse parsing tasks in two settings: \Ni when the EDUs are given (gold segmentation) and \Nii end-to-end parsing. We see that our model outperforms the baselines in both parsing conditions achieving SoTA. When gold segmentation is provided, our model outperforms the single-task training model of \citep{lin-etal-2019-unified} by 0.43\%, 1.06\% and 0.82\% absolute in Span, Nuclearity and Relation, respectively. Our parser also surpasses their joint training model, which uses multi-task  training (segmentation and parsing), with 0.61\% and 0.4\% absolute improvements in Nuclearity and Relation, respectively. For end-to-end parsing, compared to the best baseline \citep{lin-etal-2019-unified}, our model yields 0.27\%, 0.67\%, and 1.30\% absolute improvements in Span, Nuclearity, Relation, respectively. This demonstrates the effectiveness of our conditional splitting approach and end-to-end formulation of the discourse analysis task. The fact that our model improves on \emph{span} identification indicates that our method also yields better EDU segmentation.

\begin{table}[t!]
\centering
\scalebox{0.75}{\begin{tabular}{l|ccc}  
\toprule
\textbf{Approach} & \multicolumn{1}{c}{\bf{Span}} & \bf{Nuclearity} & \bf{Relation}\\
\midrule
\multicolumn{4}{c}{\textbf{Parsing with gold EDU segmentation}} \\
\bf{Human Agreement} & 95.7 & 90.4 & 83.0 \\
\midrule
\bf{Baselines} & & & \\
\citet{Wang-acl-2017} & 95.6 & 87.8 & 77.6 \\ 
\citet{lin-etal-2019-unified} (single) & 96.94 & 90.89 & 81.28 \\
\citet{lin-etal-2019-unified} (joint) & 97.44 & 91.34 & 81.70 \\ 
\bf{Our Model} & \textbf{97.37} & \textbf{91.95} & \textbf{82.10} \\
\midrule
\multicolumn{4}{c}{\textbf{End-to-End parsing}} \\
\midrule
\bf{Baselines} & & & \\
\citet{soricut-marcu-2003-sentence} & 76.7 & 70.2 & 58.0 \\
\citet{joty-etal-2012-novel} & 82.4 & 76.6 & 67.5 \\ 
\citet{lin-etal-2019-unified} (pipeline) & 91.14 & 85.80 & 76.94 \\ 
\citet{lin-etal-2019-unified} (joint) & 91.75 & 86.38  & 77.52 \\
\bf{Our Model} & \textbf{92.02} & \textbf{87.05} & \textbf{78.82} \\
\bottomrule
\end{tabular}
}
\caption{Results on discourse parsing tasks on the RST-DT test set with and without gold segmentation.}
\label{table:parseg-results}
\end{table}

\subsection{Parsing Speed Comparison}

We compare parsing speed of different  models in Table \ref{table:speed}. We ran our models on both CPU (Intel Xeon W-2133) and GPU (Nvidia GTX 1080 Ti). 

\paragraph{Syntactic Parsing}

The Berkeley Parser and ZPar are two representative non-neural parsers without access to GPUs.
\citet{minimal-span-based-parsing} employ max-margin training and perform {top-down greedy decoding} on CPUs.
Meanwhile, \citet{kitaev-klein-2018-constituency, ZhouZ19, wei-etal-2020-span} use a self-attention encoder and perform decoding using Cython for acceleration. \citet{ijcai2020-560} perform CKY decoding on GPU. The parser proposed by \citet{gomez-rodriguez-vilares-2018-constituent} is also efficient as it treats parsing as a sequence labeling task. However, its parsing accuracy is much lower compared to others (90.7 F1 in Table \ref{table:ptb_single_model}).

We see that our parser is much more efficient than existing ones. It utilizes neural modules to perform splitting, which is optimized and parallelized with efficient GPU implementation. It can parse $1,127$ sentences/second, which is faster than existing parsers. 
In fact, there is still room to improve our speed by choosing better architectures, like the Transformer which has $\mathcal{O}(1)$ running time in encoding a sentence compared to $\mathcal{O}(n)$ of the bi-LSTM encoder. Moreover, allowing tree generation by splitting the spans/nodes at the same tree level in parallel at each step can boost the speed further. We leave these extensions to future work.

\begin{table}[tb]
\centering
\scalebox{0.7}{
\begin{tabular}{lcc}  
\textbf{System} & \bf{Speed (Sents/s)} & \bf{Speedup}\\
\midrule
\bf{Syntactic Parser} & & \\
\citet{petrov-klein-2007-improved} (Berkeley)          & 6   &1.0x\\
\citet{zhu-etal-2013-fast}(ZPar)                                      & 90  &15.0x\\
\citet{minimal-span-based-parsing}                                   & 76  &12.7x\\
\citet{shen-etal-2018-straight}                                   & 111 &18.5x\\
\citet{nguyen-etal-2020-efficient}                                  & 130 &21.7x \\
\citet{ZhouZ19}                                          & 159 &26.5x\\
\citet{wei-etal-2020-span}                                          & 220 &36.7x\\
\citet{gomez-rodriguez-vilares-2018-constituent} & 780 &130x\\
\citet{kitaev-klein-2018-constituency} (GPU)                    & 830 &138.3x\\
\citet{ijcai2020-560}                                                    & 924 &154x\\
Our model (GPU)                                            & 1127 &187.3x\\
\midrule
\multicolumn{3}{l}{\bf{End-to-End Discourse parsing (Segmenter + Parser)}}  \\
CODRA \cite{joty-etal-2015-codra} & 3.05 & 1.0x \\ 
SPADE \cite{soricut-marcu-2003-sentence} & 4.90 & 1.6x \\
\cite{lin-etal-2019-unified} & 28.96 & 9.5x\\
Our end-to-end parser (CPU) & 59.03 & 19.4x\\
Our end-to-end parser (GPU) & 135.85 & 44.5x\\
\bottomrule
\end{tabular}
}
\caption{Speed comparison of our parser with existing syntactic and discourse parsers.}
\label{table:speed}
\end{table}

\paragraph{Discourse Parsing} For measuring discourse parsing speed, we follow the same set up as \citet{lin-etal-2019-unified}, and evaluate the models with the same 100 sentences randomly selected from the test set. We include the model loading time for all the systems. Since SPADE and CODRA need to extract a handful of features, they are typically slower than the neural models which use pretrained embeddings. In addition, CODRA's DCRF parser has a $O(n^3)$  inference time complexity. As shown, our parser is 4.7x faster than the fastest end-to-end parser of \citet{lin-etal-2019-unified}, making it not only effective but also highly efficient. Even when tested only on the CPU, our model is faster than all the other models which run on GPU or CPU, thanks to the end-to-end formulation that does not need EDU segmentation beforehand.

\section{Related Work}
With the recent popularity of neural architectures, such as LSTMs \citep{Hochreiter:1997} and Transformers \citep{NIPS2017_7181}, various neural models have been proposed to encode the input sentences and infer their constituency trees. To enforce structural consistency, such methods employ either a greedy transition-based \citep{dyer-etal-2016-recurrent,liu-zhang-2017-shift}, a globally optimized chart parsing \citep{gaddy-etal-2018-whats,kitaev-klein-2018-constituency}, or a greedy top-down algorithm \citep{minimal-span-based-parsing, shen-etal-2018-straight}.
Meanwhile, researchers also tried to cast the parsing problem into tasks that can be solved differently. For example, \citet{gomez-rodriguez-vilares-2018-constituent,shen-etal-2018-straight} proposed to map the syntactic tree of a sentence containing $n$ tokens into a sequence of $n-1$ labels or scalars. However, parsers of this type suffer from the exposure bias during inference. Beside these methods, Seq2Seq models have been used to generate a linearized form of the tree \citep{NIPS2015_277281aa, kamigaito-etal-2017-supervised, suzuki-etal-2018-empirical, fernandez-gonzalez-gomez-rodriguez-2020-enriched}. However, these methods may generate invalid trees when the open and end brackets do not match.

In discourse parsing, existing parsers receive the EDUs from a segmenter to build the discourse tree, which makes them susceptible to errors when the segmenter produces incorrect EDUs \citep{joty-etal-2012-novel,joty-etal-2015-codra, lin-etal-2019-unified, zhang-etal-2020-top, liu-etal-2020-multilingual-neural}. There are also attempts which model constituency and discourse parsing jointly \citep{zhao-huang-2017-joint} and do not need to perform EDU preprocessing. It is based on the finding that each EDU generally corresponds to a constituent in constituency tree, \ie\ discourse structure usually aligns with constituency structure. However, it has the drawback that it needs to build joint syntacto-discourse data set for training which is not easily adaptable to new languages and domains.

Our approach differs from previous methods in that it represents the constituency structure as a series of splitting representations, and uses a Seq2Seq framework to model the splitting decision at each step. By enabling beam search, our model can find the best trees without the need to perform an expensive global search. We also unify discourse segmentation and parsing into one system by generalizing our model, which has been done for the first time to the best of our knowledge.

Our splitting mechanism shares some similarities with Pointer Network \citep{VinyalsNIPS2015, ma-etal-2018-stack,fernandez-gonzalez-gomez-rodriguez-2019-left,fernandez-gonzalez-gomez-rodriguez-2020-transition} or head-selection approaches \citep{zhang-etal-2017-dependency-parsing, kurita-sogaard-2019-multi}, but is distinct from them that in each decoding step, our method identifies the splitting point of a span and generates a new input for future steps instead of pointing to generate the next decoder input.

\section{Conclusion}

We have presented a novel, generic parsing method for constituency parsing based on a Seq2Seq framework. Our method supports an efficient top-down decoding algorithm that uses a pointing function for scoring possible splitting points. The pointing mechanism captures global structural properties of a tree and allows efficient training with a cross entropy loss. Our formulation, when applied to discourse parsing, can bypass discourse segmentation as a pre-requisite step. Through experiments we have shown that our method outperforms all existing top-down methods on English Penn Treebank and RST Discourse Treebank sentence-level parsing tasks. With pre-trained representations, our method rivals state-of-the-art methods, while being faster. Our model also establishes a new state-of-the-art for sentence-level RST  parsing.

\bibliography{refs}

\begin{thebibliography}{52}
\expandafter\ifx\csname natexlab\endcsname\relax\def\natexlab#1{#1}\fi

\bibitem[{Artzi et~al.(2013)Artzi, FitzGerald, and
  Zettlemoyer}]{artzi-etal-2013-semantic}
Yoav Artzi, Nicholas FitzGerald, and Luke Zettlemoyer. 2013.
\newblock Semantic parsing with {C}ombinatory {C}ategorial {G}rammars.
\newblock In \emph{Proceedings of the 51st Annual Meeting of the Association
  for Computational Linguistics (Tutorials)}, page~2, Sofia, Bulgaria.
  Association for Computational Linguistics.

\bibitem[{Banarescu et~al.(2013)Banarescu, Bonial, Cai, Georgescu, Griffitt,
  Hermjakob, Knight, Koehn, Palmer, and
  Schneider}]{banarescu-etal-2013-abstract}
Laura Banarescu, Claire Bonial, Shu Cai, Madalina Georgescu, Kira Griffitt, Ulf
  Hermjakob, Kevin Knight, Philipp Koehn, Martha Palmer, and Nathan Schneider.
  2013.
\newblock {A}bstract {M}eaning {R}epresentation for sembanking.
\newblock In \emph{Proceedings of the 7th Linguistic Annotation Workshop and
  Interoperability with Discourse}, pages 178--186, Sofia, Bulgaria.
  Association for Computational Linguistics.

\bibitem[{Bengio et~al.(2015)Bengio, Vinyals, Jaitly, and
  Shazeer}]{NIPS2015_e995f98d}
Samy Bengio, Oriol Vinyals, Navdeep Jaitly, and Noam Shazeer. 2015.
\newblock \href
  {https://proceedings.neurips.cc/paper/2015/file/e995f98d56967d946471af29d7bf99f1-Paper.pdf}
  {Scheduled sampling for sequence prediction with recurrent neural networks}.
\newblock In \emph{Advances in Neural Information Processing Systems},
  volume~28, pages 1171--1179. Curran Associates, Inc.

\bibitem[{Bjorkelund et~al.(2014)Bjorkelund, Cetinoglu, Falenska, Farkas,
  Mueller, Seeker, and Szanto}]{spmrl2014}
Anders Bjorkelund, Ozlem Cetinoglu, Agnieszka Falenska, Richard Farkas, Thomas
  Mueller, Wolfgang Seeker, and Zsolt Szanto. 2014.
\newblock The ims-wrocław-szeged-cis entry at the spmrl 2014 shared task:
  Reranking and morphosyntax meet unlabeled data.
\newblock In \emph{Proceedings of the First Joint Workshop on Statistical
  Parsing of Morphologically Rich Languages and Syntactic Analysis of
  NonCanonical Languages}, pages 97--102.

\bibitem[{Coavoux and Crabb{\'e}(2017)}]{coavoux-crabbe-2017-multilingual}
Maximin Coavoux and Beno{\^\i}t Crabb{\'e}. 2017.
\newblock Multilingual lexicalized constituency parsing with word-level
  auxiliary tasks.
\newblock In \emph{Proceedings of the 15th Conference of the {E}uropean Chapter
  of the Association for Computational Linguistics: Volume 2, Short Papers},
  pages 331--336, Valencia, Spain. Association for Computational Linguistics.

\bibitem[{Cross and Huang(2016)}]{cross-huang-2016-span}
James Cross and Liang Huang. 2016.
\newblock \href {https://doi.org/10.18653/v1/D16-1001} {Span-based constituency
  parsing with a structure-label system and provably optimal dynamic oracles}.
\newblock In \emph{Proceedings of the 2016 Conference on Empirical Methods in
  Natural Language Processing}, pages 1--11, Austin, Texas. Association for
  Computational Linguistics.

\bibitem[{Devlin et~al.(2019)Devlin, Chang, Lee, and
  Toutanova}]{devlin-etal-2019-bert}
Jacob Devlin, Ming-Wei Chang, Kenton Lee, and Kristina Toutanova. 2019.
\newblock {BERT}: Pre-training of deep bidirectional transformers for language
  understanding.
\newblock In \emph{Proceedings of the 2019 Conference of the North {A}merican
  Chapter of the Association for Computational Linguistics: Human Language
  Technologies, Volume 1 (Long and Short Papers)}, pages 4171--4186,
  Minneapolis, Minnesota. Association for Computational Linguistics.

\bibitem[{Dozat and Manning(2017)}]{DozatM17}
Timothy Dozat and Christopher~D. Manning. 2017.
\newblock Deep biaffine attention for neural dependency parsing.
\newblock In \emph{5th International Conference on Learning Representations,
  {ICLR} 2017, Toulon, France, April 24-26, 2017, Conference Track
  Proceedings}.

\bibitem[{Durrett and Klein(2015)}]{durrett-klein-2015-neural}
Greg Durrett and Dan Klein. 2015.
\newblock \href {https://doi.org/10.3115/v1/P15-1030} {Neural {CRF} parsing}.
\newblock In \emph{Proceedings of the 53rd Annual Meeting of the Association
  for Computational Linguistics and the 7th International Joint Conference on
  Natural Language Processing (Volume 1: Long Papers)}, pages 302--312,
  Beijing, China. Association for Computational Linguistics.

\bibitem[{Dyer et~al.(2016)Dyer, Kuncoro, Ballesteros, and
  Smith}]{dyer-etal-2016-recurrent}
Chris Dyer, Adhiguna Kuncoro, Miguel Ballesteros, and Noah~A. Smith. 2016.
\newblock \href {https://doi.org/10.18653/v1/N16-1024} {Recurrent neural
  network grammars}.
\newblock In \emph{Proceedings of the 2016 Conference of the North {A}merican
  Chapter of the Association for Computational Linguistics: Human Language
  Technologies}, pages 199--209, San Diego, California. Association for
  Computational Linguistics.

\bibitem[{Fern{\'a}ndez-Gonz{\'a}lez and
  G{\'o}mez-Rodr{\'\i}guez(2019)}]{fernandez-gonzalez-gomez-rodriguez-2019-left}
Daniel Fern{\'a}ndez-Gonz{\'a}lez and Carlos G{\'o}mez-Rodr{\'\i}guez. 2019.
\newblock \href {https://doi.org/10.18653/v1/N19-1076} {Left-to-right
  dependency parsing with pointer networks}.
\newblock In \emph{Proceedings of the 2019 Conference of the North {A}merican
  Chapter of the Association for Computational Linguistics: Human Language
  Technologies, Volume 1 (Long and Short Papers)}, pages 710--716, Minneapolis,
  Minnesota. Association for Computational Linguistics.

\bibitem[{Fern{\'a}ndez-Gonz{\'a}lez and
  G{\'o}mez-Rodr{\'\i}guez(2020{\natexlab{a}})}]{fernandez-gonzalez-gomez-rodriguez-2020-enriched}
Daniel Fern{\'a}ndez-Gonz{\'a}lez and Carlos G{\'o}mez-Rodr{\'\i}guez.
  2020{\natexlab{a}}.
\newblock \href {https://doi.org/10.18653/v1/2020.acl-main.376} {Enriched
  in-order linearization for faster sequence-to-sequence constituent parsing}.
\newblock In \emph{Proceedings of the 58th Annual Meeting of the Association
  for Computational Linguistics}, pages 4092--4099, Online. Association for
  Computational Linguistics.

\bibitem[{Fern{\'a}ndez-Gonz{\'a}lez and
  G{\'o}mez-Rodr{\'\i}guez(2020{\natexlab{b}})}]{fernandez-gonzalez-gomez-rodriguez-2020-transition}
Daniel Fern{\'a}ndez-Gonz{\'a}lez and Carlos G{\'o}mez-Rodr{\'\i}guez.
  2020{\natexlab{b}}.
\newblock \href {https://doi.org/10.18653/v1/2020.acl-main.629}
  {Transition-based semantic dependency parsing with pointer networks}.
\newblock In \emph{Proceedings of the 58th Annual Meeting of the Association
  for Computational Linguistics}, pages 7035--7046, Online. Association for
  Computational Linguistics.

\bibitem[{Gaddy et~al.(2018)Gaddy, Stern, and Klein}]{gaddy-etal-2018-whats}
David Gaddy, Mitchell Stern, and Dan Klein. 2018.
\newblock \href {https://doi.org/10.18653/v1/N18-1091} {What{'}s going on in
  neural constituency parsers? an analysis}.
\newblock In \emph{Proceedings of the 2018 Conference of the North {A}merican
  Chapter of the Association for Computational Linguistics: Human Language
  Technologies, Volume 1 (Long Papers)}, pages 999--1010, New Orleans,
  Louisiana. Association for Computational Linguistics.

\bibitem[{G{\'o}mez and
  Vilares(2018)}]{gomez-rodriguez-vilares-2018-constituent}
Carlos G{\'o}mez, Rodr{\'\i}guez and David Vilares. 2018.
\newblock \href {https://doi.org/10.18653/v1/D18-1162} {Constituent parsing as
  sequence labeling}.
\newblock In \emph{Proceedings of the 2018 Conference on Empirical Methods in
  Natural Language Processing}, pages 1314--1324, Brussels, Belgium.
  Association for Computational Linguistics.

\bibitem[{Hochreiter and Schmidhuber(1997)}]{Hochreiter:1997}
Sepp Hochreiter and J{\"u}rgen Schmidhuber. 1997.
\newblock Long short-term memory.
\newblock \emph{Neural computation}, 9(8):1735--1780.

\bibitem[{Joty et~al.(2012)Joty, Carenini, and Ng}]{joty-etal-2012-novel}
Shafiq Joty, Giuseppe Carenini, and Raymond Ng. 2012.
\newblock A novel discriminative framework for sentence-level discourse
  analysis.
\newblock In \emph{Proceedings of the 2012 Joint Conference on Empirical
  Methods in Natural Language Processing and Computational Natural Language
  Learning}, pages 904--915, Jeju Island, Korea. Association for Computational
  Linguistics.

\bibitem[{Joty et~al.(2013)Joty, Carenini, Ng, and
  Mehdad}]{joty-etal-2013-combining}
Shafiq Joty, Giuseppe Carenini, Raymond Ng, and Yashar Mehdad. 2013.
\newblock \href {https://www.aclweb.org/anthology/P13-1048} {Combining intra-
  and multi-sentential rhetorical parsing for document-level discourse
  analysis}.
\newblock In \emph{Proceedings of the 51st Annual Meeting of the Association
  for Computational Linguistics (Volume 1: Long Papers)}, pages 486--496,
  Sofia, Bulgaria. Association for Computational Linguistics.

\bibitem[{Joty et~al.(2015)Joty, Carenini, and Ng}]{joty-etal-2015-codra}
Shafiq Joty, Giuseppe Carenini, and Raymond~T. Ng. 2015.
\newblock {CODRA}: A novel discriminative framework for rhetorical analysis.
\newblock \emph{Computational Linguistics}, 41(3):385--435.

\bibitem[{Kamigaito et~al.(2017)Kamigaito, Hayashi, Hirao, Takamura, Okumura,
  and Nagata}]{kamigaito-etal-2017-supervised}
Hidetaka Kamigaito, Katsuhiko Hayashi, Tsutomu Hirao, Hiroya Takamura, Manabu
  Okumura, and Masaaki Nagata. 2017.
\newblock \href {https://www.aclweb.org/anthology/I17-2002} {Supervised
  attention for sequence-to-sequence constituency parsing}.
\newblock In \emph{Proceedings of the Eighth International Joint Conference on
  Natural Language Processing (Volume 2: Short Papers)}, pages 7--12, Taipei,
  Taiwan. Asian Federation of Natural Language Processing.

\bibitem[{Kingma and Ba(2015)}]{KingmaB14}
Diederik~P. Kingma and Jimmy Ba. 2015.
\newblock Adam: {A} method for stochastic optimization.
\newblock In \emph{3rd International Conference on Learning Representations,
  {ICLR} 2015, San Diego, CA, USA, May 7-9, 2015, Conference Track
  Proceedings}.

\bibitem[{Kitaev et~al.(2019)Kitaev, Cao, and
  Klein}]{kitaev-etal-2019-multilingual}
Nikita Kitaev, Steven Cao, and Dan Klein. 2019.
\newblock \href {https://doi.org/10.18653/v1/P19-1340} {Multilingual
  constituency parsing with self-attention and pre-training}.
\newblock In \emph{Proceedings of the 57th Annual Meeting of the Association
  for Computational Linguistics}, pages 3499--3505, Florence, Italy.
  Association for Computational Linguistics.

\bibitem[{Kitaev and Klein(2018)}]{kitaev-klein-2018-constituency}
Nikita Kitaev and Dan Klein. 2018.
\newblock \href {https://doi.org/10.18653/v1/P18-1249} {Constituency parsing
  with a self-attentive encoder}.
\newblock In \emph{Proceedings of the 56th Annual Meeting of the Association
  for Computational Linguistics (Volume 1: Long Papers)}, pages 2676--2686,
  Melbourne, Australia. Association for Computational Linguistics.

\bibitem[{Kurita and S{\o}gaard(2019)}]{kurita-sogaard-2019-multi}
Shuhei Kurita and Anders S{\o}gaard. 2019.
\newblock \href {https://doi.org/10.18653/v1/P19-1232} {Multi-task semantic
  dependency parsing with policy gradient for learning easy-first strategies}.
\newblock In \emph{Proceedings of the 57th Annual Meeting of the Association
  for Computational Linguistics}, pages 2420--2430, Florence, Italy.
  Association for Computational Linguistics.

\bibitem[{Lin et~al.(2019)Lin, Joty, Jwalapuram, and
  Bari}]{lin-etal-2019-unified}
Xiang Lin, Shafiq Joty, Prathyusha Jwalapuram, and M~Saiful Bari. 2019.
\newblock \href {https://doi.org/10.18653/v1/P19-1410} {A unified linear-time
  framework for sentence-level discourse parsing}.
\newblock In \emph{Proceedings of the 57th Annual Meeting of the Association
  for Computational Linguistics}, pages 4190--4200, Florence, Italy.
  Association for Computational Linguistics.

\bibitem[{Liu and Zhang(2017)}]{liu-zhang-2017-shift}
Jiangming Liu and Yue Zhang. 2017.
\newblock Shift-reduce constituent parsing with neural lookahead features.
\newblock \emph{Transactions of the Association for Computational Linguistics},
  5:45--58.

\bibitem[{Liu et~al.(2020)Liu, Shi, and
  Chen}]{liu-etal-2020-multilingual-neural}
Zhengyuan Liu, Ke~Shi, and Nancy Chen. 2020.
\newblock \href {https://doi.org/10.18653/v1/2020.coling-main.591}
  {Multilingual neural {RST} discourse parsing}.
\newblock In \emph{Proceedings of the 28th International Conference on
  Computational Linguistics}, pages 6730--6738, Barcelona, Spain (Online).
  International Committee on Computational Linguistics.

\bibitem[{Lynn et~al.(2002)Lynn, Marcu, and Okurowski.}]{RST:DT:2002}
Carlson Lynn, Daniel Marcu, and Mary~Ellen Okurowski. 2002.
\newblock Rst discourse treebank (rst--dt) ldc2002t07.
\newblock \emph{Linguistic Data Consortium}.

\bibitem[{Ma et~al.(2018)Ma, Hu, Liu, Peng, Neubig, and
  Hovy}]{ma-etal-2018-stack}
Xuezhe Ma, Zecong Hu, Jingzhou Liu, Nanyun Peng, Graham Neubig, and Eduard
  Hovy. 2018.
\newblock \href {https://doi.org/10.18653/v1/P18-1130} {Stack-pointer networks
  for dependency parsing}.
\newblock In \emph{Proceedings of the 56th Annual Meeting of the Association
  for Computational Linguistics (Volume 1: Long Papers)}, pages 1403--1414,
  Melbourne, Australia. Association for Computational Linguistics.

\bibitem[{Mann and Thompson(1988)}]{Mann88}
William Mann and Sandra Thompson. 1988.
\newblock {Rhetorical Structure Theory: Toward a Functional Theory of Text
  Organization}.
\newblock \emph{Text}, 8(3):243--281.

\bibitem[{Marcus et~al.(1993)Marcus, Marcinkiewicz, and
  Santorini}]{PTB:Marcus:1993}
Mitchell~P. Marcus, Mary~Ann Marcinkiewicz, and Beatrice Santorini. 1993.
\newblock Building a large annotated corpus of english: The penn treebank.
\newblock \emph{Comput. Linguist.}, 19(2):313--330.

\bibitem[{Nguyen et~al.(2020)Nguyen, Nguyen, Joty, and
  Li}]{nguyen-etal-2020-efficient}
Thanh-Tung Nguyen, Xuan-Phi Nguyen, Shafiq Joty, and Xiaoli Li. 2020.
\newblock Efficient constituency parsing by pointing.
\newblock In \emph{Proceedings of the 58th Annual Meeting of the Association
  for Computational Linguistics}, pages 3284--3294, Online. Association for
  Computational Linguistics.

\bibitem[{Peters et~al.(2018)Peters, Neumann, Iyyer, Gardner, Clark, Lee, and
  Zettlemoyer}]{Peters:2018}
Matthew~E. Peters, Mark Neumann, Mohit Iyyer, Matt Gardner, Christopher Clark,
  Kenton Lee, and Luke Zettlemoyer. 2018.
\newblock Deep contextualized word representations.
\newblock In \emph{Proc. of NAACL}.

\bibitem[{Petrov and Klein(2007)}]{petrov-klein-2007-improved}
Slav Petrov and Dan Klein. 2007.
\newblock Improved inference for unlexicalized parsing.
\newblock In \emph{Human Language Technologies 2007: The Conference of the
  North {A}merican Chapter of the Association for Computational Linguistics;
  Proceedings of the Main Conference}, pages 404--411, Rochester, New York.
  Association for Computational Linguistics.

\bibitem[{Seddah et~al.(2013)Seddah, Tsarfaty, K{\"u}bler, Candito, Choi,
  Farkas, Foster, Goenaga, Gojenola~Galletebeitia, Goldberg, Green, Habash,
  Kuhlmann, Maier, Nivre, Przepi{\'o}rkowski, Roth, Seeker, Versley, Vincze,
  Woli{\'n}ski, Wr{\'o}blewska, and Villemonte de~la
  Clergerie}]{seddah-etal-2013-overview}
Djam{\'e} Seddah, Reut Tsarfaty, Sandra K{\"u}bler, Marie Candito, Jinho~D.
  Choi, Rich{\'a}rd Farkas, Jennifer Foster, Iakes Goenaga, Koldo
  Gojenola~Galletebeitia, Yoav Goldberg, Spence Green, Nizar Habash, Marco
  Kuhlmann, Wolfgang Maier, Joakim Nivre, Adam Przepi{\'o}rkowski, Ryan Roth,
  Wolfgang Seeker, Yannick Versley, Veronika Vincze, Marcin Woli{\'n}ski, Alina
  Wr{\'o}blewska, and Eric Villemonte de~la Clergerie. 2013.
\newblock Overview of the {SPMRL} 2013 shared task: A cross-framework
  evaluation of parsing morphologically rich languages.
\newblock In \emph{Proceedings of the Fourth Workshop on Statistical Parsing of
  Morphologically-Rich Languages}, pages 146--182, Seattle, Washington, USA.
  Association for Computational Linguistics.

\bibitem[{Shen et~al.(2018)Shen, Lin, Jacob, Sordoni, Courville, and
  Bengio}]{shen-etal-2018-straight}
Yikang Shen, Zhouhan Lin, Athul~Paul Jacob, Alessandro Sordoni, Aaron
  Courville, and Yoshua Bengio. 2018.
\newblock Straight to the tree: Constituency parsing with neural syntactic
  distance.
\newblock In \emph{Proceedings of the 56th Annual Meeting of the Association
  for Computational Linguistics (Volume 1: Long Papers)}, pages 1171--1180,
  Melbourne, Australia. Association for Computational Linguistics.

\bibitem[{Soricut and Marcu(2003)}]{soricut-marcu-2003-sentence}
Radu Soricut and Daniel Marcu. 2003.
\newblock Sentence level discourse parsing using syntactic and lexical
  information.
\newblock In \emph{Proceedings of the 2003 Human Language Technology Conference
  of the North {A}merican Chapter of the Association for Computational
  Linguistics}, pages 228--235.

\bibitem[{Stern et~al.(2017{\natexlab{a}})Stern, Andreas, and
  Klein}]{minimal-span-based-parsing}
Mitchell Stern, Jacob Andreas, and Dan Klein. 2017{\natexlab{a}}.
\newblock A minimal span-based neural constituency parser.
\newblock In \emph{Proceedings of the 55th Annual Meeting of the Association
  for Computational Linguistics, {ACL} 2017, Vancouver, Canada, July 30 -
  August 4, Volume 1: Long Papers}, pages 818--827.

\bibitem[{Stern et~al.(2017{\natexlab{b}})Stern, Fried, and
  Klein}]{stern-etal-2017-effective}
Mitchell Stern, Daniel Fried, and Dan Klein. 2017{\natexlab{b}}.
\newblock \href {https://doi.org/10.18653/v1/D17-1178} {Effective inference for
  generative neural parsing}.
\newblock In \emph{Proceedings of the 2017 Conference on Empirical Methods in
  Natural Language Processing}, pages 1695--1700, Copenhagen, Denmark.
  Association for Computational Linguistics.

\bibitem[{Suzuki et~al.(2018)Suzuki, Takase, Kamigaito, Morishita, and
  Nagata}]{suzuki-etal-2018-empirical}
Jun Suzuki, Sho Takase, Hidetaka Kamigaito, Makoto Morishita, and Masaaki
  Nagata. 2018.
\newblock \href {https://doi.org/10.18653/v1/P18-2097} {An empirical study of
  building a strong baseline for constituency parsing}.
\newblock In \emph{Proceedings of the 56th Annual Meeting of the Association
  for Computational Linguistics (Volume 2: Short Papers)}, pages 612--618,
  Melbourne, Australia. Association for Computational Linguistics.

\bibitem[{Vaswani et~al.(2017)Vaswani, Shazeer, Parmar, Uszkoreit, Jones,
  Gomez, Kaiser, and Polosukhin}]{NIPS2017_7181}
Ashish Vaswani, Noam Shazeer, Niki Parmar, Jakob Uszkoreit, Llion Jones,
  Aidan~N Gomez, \L~ukasz Kaiser, and Illia Polosukhin. 2017.
\newblock Attention is all you need.
\newblock In I.~Guyon, U.~V. Luxburg, S.~Bengio, H.~Wallach, R.~Fergus,
  S.~Vishwanathan, and R.~Garnett, editors, \emph{Advances in Neural
  Information Processing Systems 30}, pages 5998--6008. Curran Associates, Inc.

\bibitem[{Vinyals et~al.(2015{\natexlab{a}})Vinyals, Fortunato, and
  Jaitly}]{VinyalsNIPS2015}
Oriol Vinyals, Meire Fortunato, and Navdeep Jaitly. 2015{\natexlab{a}}.
\newblock Pointer networks.
\newblock In C.~Cortes, N.~D. Lawrence, D.~D. Lee, M.~Sugiyama, and R.~Garnett,
  editors, \emph{Advances in Neural Information Processing Systems 28}, pages
  2692--2700. Curran Associates, Inc.

\bibitem[{Vinyals et~al.(2015{\natexlab{b}})Vinyals, Kaiser, Koo, Petrov,
  Sutskever, and Hinton}]{NIPS2015_277281aa}
Oriol Vinyals, \L~ukasz Kaiser, Terry Koo, Slav Petrov, Ilya Sutskever, and
  Geoffrey Hinton. 2015{\natexlab{b}}.
\newblock \href
  {https://proceedings.neurips.cc/paper/2015/file/277281aada22045c03945dcb2ca6f2ec-Paper.pdf}
  {Grammar as a foreign language}.
\newblock In \emph{Advances in Neural Information Processing Systems},
  volume~28, pages 2773--2781. Curran Associates, Inc.

\bibitem[{Wang et~al.(2017)Wang, Li, and Wang}]{Wang-acl-2017}
Yizhong Wang, Sujian Li, and Houfeng Wang. 2017.
\newblock A two-stage parsing method for text-level discourse analysis.
\newblock In \emph{Proceedings of the 55th Annual Meeting of the Association
  for Computational Linguistics (Volume 2: Short Papers)}, pages 184--188.
  Association for Computational Linguistics.

\bibitem[{Webber(2004)}]{Webber04}
B.~Webber. 2004.
\newblock {D-LTAG: Extending Lexicalized TAG to Discourse}.
\newblock \emph{Cognitive Science}, 28(5):751--779.

\bibitem[{Wei et~al.(2020)Wei, Wu, and Lan}]{wei-etal-2020-span}
Yang Wei, Yuanbin Wu, and Man Lan. 2020.
\newblock A span-based linearization for constituent trees.
\newblock In \emph{Proceedings of the 58th Annual Meeting of the Association
  for Computational Linguistics}, pages 3267--3277, Online. Association for
  Computational Linguistics.

\bibitem[{Zhang et~al.(2020{\natexlab{a}})Zhang, Xing, Kong, Li, and
  Zhou}]{zhang-etal-2020-top}
Longyin Zhang, Yuqing Xing, Fang Kong, Peifeng Li, and Guodong Zhou.
  2020{\natexlab{a}}.
\newblock \href {https://doi.org/10.18653/v1/2020.acl-main.569} {A top-down
  neural architecture towards text-level parsing of discourse rhetorical
  structure}.
\newblock In \emph{Proceedings of the 58th Annual Meeting of the Association
  for Computational Linguistics}, pages 6386--6395, Online. Association for
  Computational Linguistics.

\bibitem[{Zhang et~al.(2017)Zhang, Cheng, and
  Lapata}]{zhang-etal-2017-dependency-parsing}
Xingxing Zhang, Jianpeng Cheng, and Mirella Lapata. 2017.
\newblock \href {https://www.aclweb.org/anthology/E17-1063} {Dependency parsing
  as head selection}.
\newblock In \emph{Proceedings of the 15th Conference of the {E}uropean Chapter
  of the Association for Computational Linguistics: Volume 1, Long Papers},
  pages 665--676, Valencia, Spain. Association for Computational Linguistics.

\bibitem[{Zhang et~al.(2020{\natexlab{b}})Zhang, Zhou, and Li}]{ijcai2020-560}
Yu~Zhang, Houquan Zhou, and Zhenghua Li. 2020{\natexlab{b}}.
\newblock \href {https://doi.org/10.24963/ijcai.2020/560} {Fast and accurate
  neural crf constituency parsing}.
\newblock In \emph{Proceedings of the Twenty-Ninth International Joint
  Conference on Artificial Intelligence, {IJCAI-20}}, pages 4046--4053.
  International Joint Conferences on Artificial Intelligence Organization.
\newblock Main track.

\bibitem[{Zhao and Huang(2017)}]{zhao-huang-2017-joint}
Kai Zhao and Liang Huang. 2017.
\newblock \href {https://doi.org/10.18653/v1/D17-1225} {Joint
  syntacto-discourse parsing and the syntacto-discourse treebank}.
\newblock In \emph{Proceedings of the 2017 Conference on Empirical Methods in
  Natural Language Processing}, pages 2117--2123, Copenhagen, Denmark.
  Association for Computational Linguistics.

\bibitem[{Zhou and Zhao(2019)}]{ZhouZ19}
Junru Zhou and Hai Zhao. 2019.
\newblock Head-driven phrase structure grammar parsing on penn treebank.
\newblock In \emph{Proceedings of the 57th Conference of the Association for
  Computational Linguistics, {ACL} 2019, Florence, Italy, July 28- August 2,
  2019, Volume 1: Long Papers}, pages 2396--2408.

\bibitem[{Zhu et~al.(2013)Zhu, Zhang, Chen, Zhang, and
  Zhu}]{zhu-etal-2013-fast}
Muhua Zhu, Yue Zhang, Wenliang Chen, Min Zhang, and Jingbo Zhu. 2013.
\newblock Fast and accurate shift-reduce constituent parsing.
\newblock In \emph{Proceedings of the 51st Annual Meeting of the Association
  for Computational Linguistics (Volume 1: Long Papers)}, pages 434--443,
  Sofia, Bulgaria. Association for Computational Linguistics.

\end{thebibliography}
\bibliographystyle{acl_natbib}

\section*{Appendix}

\subsection{Discourse Parsing Architecture}
\begin{figure*}[t!]
\centering
\includegraphics[width=0.85\textwidth]{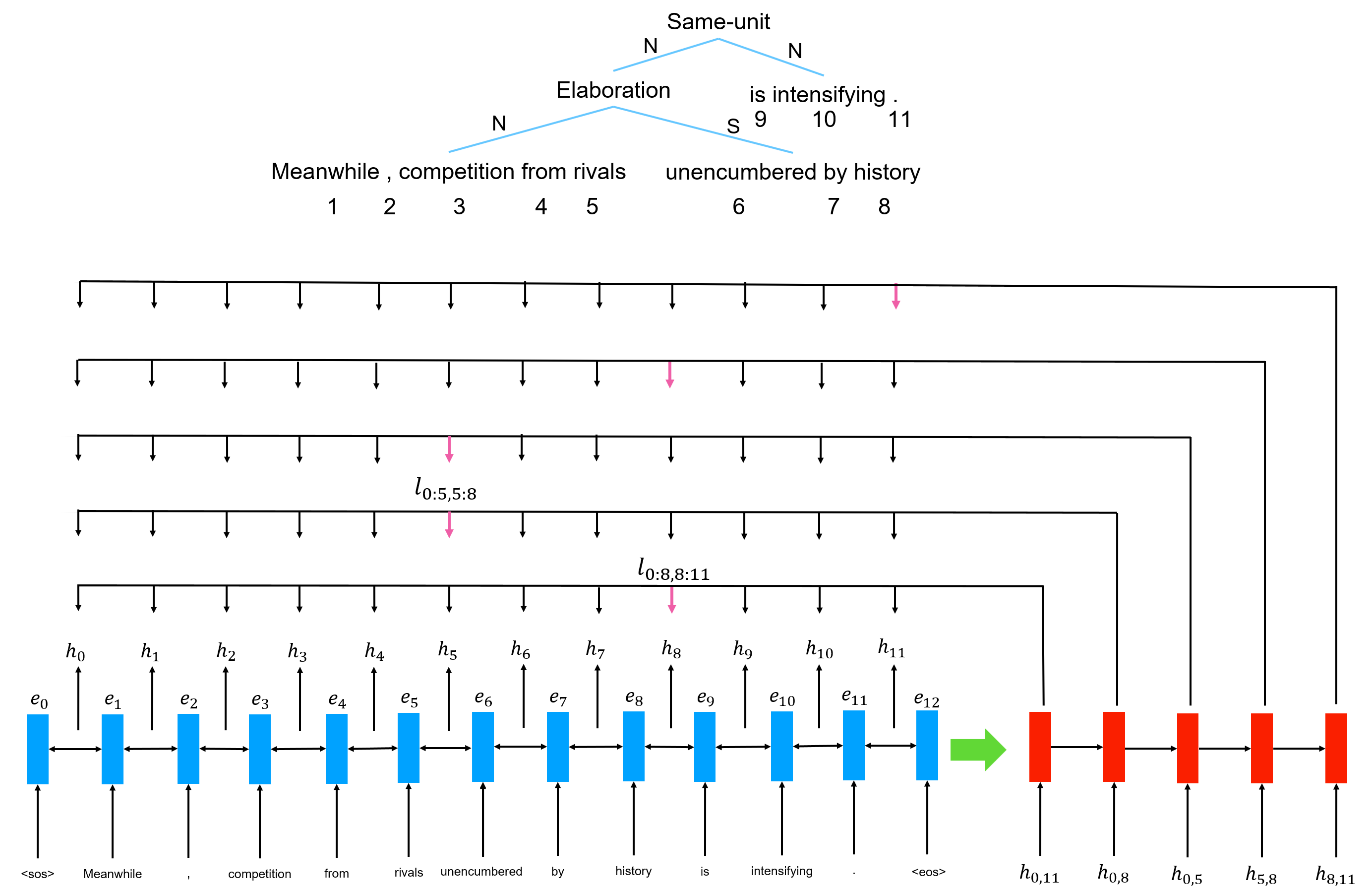}
\caption{Our discourse parser a long with the decoding process for a given sentence. The input to the decoder at each step is the representation of the span to be split. We predict splitting point using the biaffine function between the corresponding decoder state and the boundary representations. The relationship between left and right spans are assigned with the label using the label classifier. }
\label{fig:cdiscourse_parsing_whole_architecture}
\end{figure*}

Figure \ref{fig:cdiscourse_parsing_whole_architecture} illustrates our end-to-end model architecture for discourse parsing.

\subsection{Discourse Parsing Inference Algorithms}
Algorithm \ref{alg2:discourse-parsing} shows the end-to-end discourse parsing inference process.
\begin{algorithm}[h!]]
\small
    \caption{Discourse Inference}
    \label{alg2:discourse-parsing}
    \begin{algorithmic}
    \REQUIRE Sentence length $n$; boundary encoder states: $(h_0, h_1, \ldots, h_n)$; label scores: $P(l|(i,k), (k,j))$, $0\leq i < k \leq j \leq n, l \in L$, initial decoder state $st$.
    \ENSURE Parse tree $\gT$
    \STATE $\gST=[(1,n)]$  \algorithmiccomment{stack of spans}
    \STATE $\gS=[]$
    \WHILE {$\gST \neq \varnothing$}
        \STATE $(i, j) = \pop(\gST)$
        \STATE $prob, st=dec(st, (i,j))$
        \STATE $k=\argmax_{i<k \leq j}prob $
            \STATE $\text{curr\_partial\_tree}=partial\_tree$
                \IF{$j-1 > k > i+1$}    
                    \STATE $\push(\gST,(k,j))$
                    \STATE $\push(\gST,(i,k))$
                \ELSIF{$j-1 > k = i+1 $}    
                    \STATE $\push(\gST,(k,j))$
                \ELSIF{$k=j-1 > i+1$}    
                    \STATE $\push(\gST,(i,k))$
                \ENDIF
                \IF{$k \neq j$}    
                    \STATE $\push(\gS((i,k,j))$
                \ENDIF
    \ENDWHILE
    \STATE $\gT=[((i,k,j),argmax_{l}P(l|(i,k)(k,j)) \forall (i,k,j) \in \gS]$
  \end{algorithmic}
\end{algorithm}


\end{document}